\pdfoutput=1

\documentclass[11pt]{article}

\usepackage{acl}

\usepackage{times}
\usepackage{latexsym}

\usepackage[T1]{fontenc}

\usepackage[utf8]{inputenc}

\usepackage{microtype}
\usepackage{times}
\usepackage{latexsym}
\usepackage{booktabs} 
\usepackage{multirow}
\usepackage{url}

\usepackage{amsmath}
\usepackage{bm}
\usepackage{amsfonts}
\usepackage{graphicx}
\usepackage{diagbox}
\usepackage{xspace}
\usepackage{courier}
\usepackage{mathrsfs}
\usepackage{appendix}
\usepackage{float}
\usepackage{framed} 
\usepackage{amsfonts}
\usepackage{amssymb}
\usepackage{diagbox}
\usepackage{microtype}
\usepackage{textcomp}
\usepackage{amsfonts}
\usepackage{amssymb}
\usepackage{diagbox}
\usepackage{algorithm}
\usepackage{algorithmic}
\usepackage{color}
\newcommand\blarge{$_{\small \texttt{L}}$\xspace}

\newcommand\Ourmodel{UNTER\xspace}

\newcommand{\femph}[1]{\cellcolor[HTML]{C5E0B4}{#1}}
\newcommand{\semph}[1]{\cellcolor[HTML]{CFE2F3}{#1}}

\title{ \Ourmodel:  A Unified Knowledge Interface for Enhancing Pre-trained Language Models}

 \author{Deming Ye$^{1}$, Yankai Lin$^{2}$, Zhengyan Zhang$^{1}$,   Maosong Sun$^{1}$\thanks{ \ \ Corresponding author: M. Sun (sms@tsinghua.edu.cn)}\\
$^1$Department of Computer Science and Technology, Tsinghua University, Beijing, China\\
Institute for Artificial Intelligence, Tsinghua University, Beijing, China\\
Beijing National Research Center for Information Science and Technology\\
$^{2}$Gaoling School of Artificial Intelligence, Renmin University of China, Beijing\\
\texttt{yedeming001@163.com}
}

\begin{document}
\maketitle
\begin{abstract}
Recent research demonstrates that external knowledge injection can advance pre-trained language models (PLMs) in a variety of downstream NLP tasks. 
However, existing knowledge injection methods are either applicable to structured knowledge or unstructured knowledge, lacking a unified usage. 
In this paper, we propose a \textbf{UN}ified knowledge in\textbf{TER}face, \textbf{\Ourmodel}, to provide a unified perspective to exploit both structured knowledge and unstructured knowledge. In \Ourmodel, we adopt the decoder as a unified knowledge interface, aligning span representations obtained from the encoder with their corresponding knowledge. 
This approach enables the encoder to uniformly invoke span-related knowledge from its parameters for downstream applications.  
Experimental results show that, with both forms of knowledge injected, \Ourmodel gains continuous improvements on a series of knowledge-driven NLP tasks, including entity typing, named entity recognition and relation extraction, especially in low-resource scenarios.  
\end{abstract}



\section{Introduction}
 
Recent years have witnessed rapid development in enhancing pre-trained language models (PLMs).  With various kinds of knowledge injected, PLMs have achieved dramatic success in a wide range of tasks, including language modeling~\cite{LAMA, ye2022pelt}, question answering~\cite{REALM, RAGplus}, information extraction~\cite{ERNIE, KEPLER}, and syntactic parsing~\cite{LinguisticsBERT}.

Generally, knowledge used to enhance PLMs is divided into two categories, structured knowledge and unstructured knowledge~\cite{wikidata}, which have significant differences in the knowledge injection process. For example, models with structured knowledge injected ~\cite{ERNIE,KnowBERT} usually first transform the structured knowledge into knowledge embedding, followed by applying an entity linker to locate entity position and using additional modules to integrate the knowledge embedding and the original token embedding. In contrast, models with unstructured knowledge injected tend to employ a retriever to obtain relevant text and directly append it to the input text as a complement~\cite{REALM, RAGplus}. 
Due to their huge differences in the knowledge injection process,  existing knowledge-enhanced models mainly focus on either structured knowledge~\cite{LinguisticsBERT, K-BERT, commonsebert} or unstructured knowledge~\cite{TEK,dictbert}, resulting in a lack of a unified perspective to take full advantage of both forms of knowledge. 

\begin{table}[!t]
\centering
\small
\begin{tabular}{p{7.5cm}}
\toprule
\emph{Unstructured Wikipedia page of Steve Jobs}:  \\
... \textbf{Jobs} attended \textbf{Reed College} in 1972 before withdrawing that same year. ... \\
\midrule
\emph{Structured factual knowledge with aligned text}: \\
\emph{Fact}: (Ayaan Ali, educated at, Leiden University)\\
\emph{Aligned Text:}  He joined the \textbf{Leiden University} as an assistant professor, where he taught students such as the later member of parliament \textbf{Ayaan Ali}. \\
\midrule
\emph{RE Example}: He was a professor at \textbf{Reed College}, where he taught \textbf{Steve Jobs}. \\
\emph{Prediction}:  (Steve Jobs, educated at, Reed College)\\

\bottomrule
\end{tabular}
\caption{An illustration of how the knowledge injection enhances relation extraction (RE). Our model absorbs both unstructured and structured knowledge to make the proper prediction.}
\label{tab:case_study}
\end{table}


In this work, we explore the \emph{unified knowledge injection}. Instead of applying different components to various forms of knowledge, we adopt a unified interface to inject structured and unstructured knowledge into the model's parameters.  
After injection, the enhanced model can invoke the injected knowledge from its parameters in a holistic manner, for which it can be flexibly used as a vanilla PLM in downstream applications without specific plugins, such as entity linker~\cite{tagme} or retriever~\cite{DPR}.


Towards unified knowledge injection, we propose a  \textbf{UN}ified knowledge in\textbf{TER}face,  \textbf{\Ourmodel}, to provide a unified perspective to inject knowledge into the pre-trained encoder model. We adopt an encoder-decoder framework and require the decoder to generate span-related knowledge based on the span representations obtained from the encoder. Through the decoder interface, the span-related knowledge will be transferred to gradient supervision to optimize the encoder representation, which aligns the span representation to their corresponding knowledge, enabling downstream models to invoke span-related knowledge from the enhanced representation. 
Notably, we restrict the capacity of the decoder, e.g., using a shallow decoder~\cite{Weakdecoder, MAE}, to decrease the dependencies among generated tokens as well as improve the correlation between span representations and their associated knowledge. 

We construct the pre-training corpus using Wikipedia annotations and Wikidata facts. As shown in Table~\ref{tab:case_study}, we adopt two kinds of knowledge. (1) \emph{Unstructured}: We use the Wikipedia page of the anchor span to enrich the background knowledge. (2) \emph{Structured}: We align the relational fact with sentences in which it appears to obtain the demonstration of the relation.  
After absorbing these two forms of  knowledge,  our model correctly predict the factual triple, (Steve Jobs, educated at, Reed College), from the given sentence.

We conduct experiments on a series of knowledge-driven NLP tasks, including entity typing, named entity recognition and relation classification. Experimental results show that, with unstructured entity page injection,  \Ourmodel (E) achieves better performance on all evaluation tasks, especially on the entity-aware tasks. With the structured factual knowledge injection, \Ourmodel (R) gains significant enhancements on the relation-aware tasks.  
By absorbing both forms of knowledge, \Ourmodel (E+R) accomplishes continuous improvements on both entity-aware and relation-aware tasks, which demonstrates that our method can effectively enhance PLM by injecting divergent forms of knowledge.


\section{Related Work}

In this section, we introduce several common injection methods respectively designed for the structured and unstructured knowledge. 

\paragraph{Structured Knowledge} PLMs have been shown to be enhanced by injecting various structured knowledge such as factual knowledge~\cite{ERNIE}, linguistic knowledge~\cite{LinguisticsBERT,LiBERT,sensebert}, and commonsense knowledge~\cite{commonsebert}. Among them, the factual knowledge has been proven to aid PLM in the broadest range of NLP tasks~\cite{K-Adapter,KEPLER}.  
To take good advantage of related factual knowledge, a straightforward approach is to apply an external entity linker~\cite{tagme} to align entities in context with the knowledge graph. Given the linked entities, ERNIE~\cite{ERNIE} and KnowBERT~\cite{KnowBERT} adopt additional Transformer layers to integrate the original token embeddings and external knowledge embeddings. K-BERT~\cite{K-BERT} and LUKE~\cite{LUKE} directly insert mapped entity embeddings into the input sequence to provide background knowledge.  In addition, researchers also infuse the linking information into model parameters by supervising the anchor text with knowledge-aware objectives, such as replaced entity detection~\cite{WKLM},  relation prediction~\cite{K-Adapter}, and knowledge graph completion~\cite{CoLAKE}, which enables models to recall knowledge without the help of extra entity linkers. Despite the success in a wide range of tasks, existing  specific designs for structured knowledge cannot be directly applied to unstructured knowledge for unified modeling.  





\paragraph{Unstructured Knowledge}  Recent unstructured knowledge injection methods tend to append related text to the input text, followed by employing the attention mechanism to utilize the supplementary information. 
According to the knowledge source used, injection methods can be roughly divided into the following categories. 
(1) \textbf{Entity-Related Text}: TEK~\cite{TEK}, E-BERT~\cite{LAMA}, and  PELT~\cite{ye2022pelt} locate entities in the paragraphs and then insert entity pages or entity embeddings, obtained from sentences where the entity occurs, into the input.
(2) \textbf{Retrieved Text}:  REALM~\cite{REALM} and RAG~\cite{RAGplus}  retrieve question-related articles from a large corpus as background knowledge.   
RETRO~\cite{RETRO}  scales the corpus to  trillions tokens and applies a chunked cross-attention for acceleration.  
(3) \textbf{Demonstration}: REINA~\cite{REINA} and  RETROPROMPT~\cite{RETROPROMPT} retrieve similar demonstrations from the training dataset to aid  model's prediction on full-data setting and few-shot setting respectively.  
(4) \textbf{Dictionary}:    Dict-BERT~\cite{dictbert} align the rare word representation to their dictionary description to enhance the general understanding of PLMs. 
Despite great progress on a wide variety of NLP tasks, especially in question answering, existing methods for unstructured knowledge injection are usually time-consuming due to the long concatenation of texts. 






\begin{figure*}
    \centering
    \includegraphics[width=\linewidth]{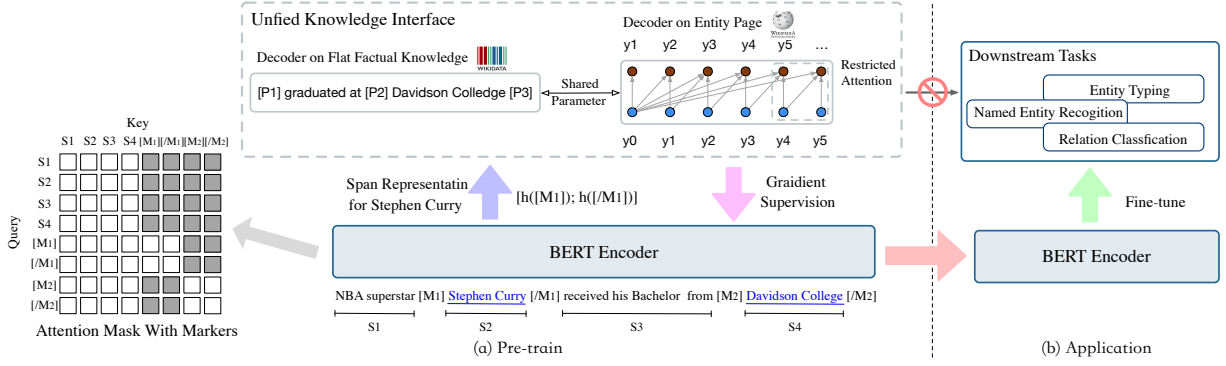}
    \caption{
     A visualized explanation on how \Ourmodel works. In the pre-training process, we first use the levitated marker with restricted attention to obtain span representations. Given a span representation, we require a shared decoder to generate span's  structured factual knowledge and unstructured Wikipedia page, which provides encoder with gradient supervision. Particularly, for structured knowledge, we flatten them by inserting soft prompts; for unstructured knowledge, we restrict the visibility of generated tokens to previous tokens.
     After the pre-training phase, we discard the decoder and only use the encoder in downstream applications.  Due to the space limit, we omit the tail entity’s  decoding process for its description in the figure.
    }
    \label{fig:model}
\end{figure*}
\section{Method}

In this section, we first introduce the approach to obtaining the span representation from the encoder, followed by the method for supervising the span representation through a unified knowledge interface.  
Then, we present the training objective and the construction of training data in detail. 


\subsection{Marker-Based Span Representation}
PLMs conduct pre-training tasks on the tokens' contextual representations~\cite{BERT}. 
In \Ourmodel, we retain the original pre-training task  to preserve the model's capabilities in general understanding.  
To avoid interfering with the original training objective on token representations, we adopt the levitated marker~\cite{PURE,ye2022plmarker} to acquire span representations from the encoder, which is also beneficial for subsequent knowledge injection.   

Formally, given an input sentence and a span $s$,  we insert a pair of text markers,  \texttt{[M]} and \texttt{[/M]}, into the input sentence to highlight the given span. To locate a span, markers are set to share the same position id with the span's boundary tokens.  


Moreover, as shown in Figure~\ref{fig:model},  to support multi-span computation, a pair of markers are tied by a directional attention.  The markers within a pair are set to be visible to each other in the attention mask matrix, but not to the text token and other pairs of markers. Going through PLMs' layers,  markers pay attention to text tokens for information aggregation. 
 
We convey the input sequence with markers to the PLM  to obtain  markers' contextual representation, $\bm{h}_{\texttt{[M]}}$ and $\bm{h}_{\texttt{[/M]}}$. Then, we concatenate them as the representation of their associated span $s$: 
\begin{equation}
    \bm{g}_s = [\bm{h}_{\texttt{[M]}}; \bm{h}_{\texttt{[/M]}}],
\end{equation}
where $[\,;\,]$ denotes the concatenation operation.


\subsection{Unified Knowledge Interface}
To alleviate the differences in the structured and unstructured knowledge injection process,  
we unify all forms of knowledge into the same semantic space by using soft prompts~\cite{PPTuning, PTR}. 
After that, we require the decoder to generate the span-related knowledge based on the span representation from encoder. 
Through the decoder interface, the span-related knowledge is  transferred to gradient supervision to optimize the span representation, which enables downstream models to recall the knowledge from the enhanced span representation. 


Notably, the token generation process in the decoder depends on parameters of both encoder and decoder. To decrease the dependency among the generated tokens and improve the correlation between span representations and their associated knowledge,  we limit the capacity of the decoder so as to store as much injected knowledge as possible to the encoder during injection. Specifically, we employ a shallow decoder with fewer layers in our encoder-decoder framework. 

Moreover, according to different knowledge forms,  we customize some features for structured and unstructured knowledge respectively. 

\paragraph{Structured Knowledge} We insert several soft prompts~\cite{PPTuning, PTR} to convert structured knowledge into coherent text. For example, the structured factual triple of the head entry \emph{Stephen Curry},  

({Stephen Curry}, {graduated at}, {Davidson College})

\noindent
is conveyed to its textual property, 

[P1] graduated at [P2] Davidson College [P3],

\noindent
where [P]s refer to trainable soft prompts. 

During the training process, the soft prompts automatically learn to make the sentence coherent in semantic space. Since the converted sentence length is relatively short, we simply use a sequential attention decoder to emphasize the dependency of tail entry on the relation. Formally, the structured decoder loss for span $s$ is defined as:

\begin{equation}
\begin{aligned}
 &\mathcal{L}_{\text{struct}}(Y | s, \theta_{dec}) = \\
 & -\sum_{t=1}^m\log P(y_t|y_{<t}, \bm{g}_s, \theta_{dec}),
\end{aligned}
\label{eq:strucut}
\end{equation} 
where  $y_{<t}$ are all previous tokens before $y_t$ and $\theta_{dec}$ is the parameter of the decoder.

\paragraph{Unstructured knowledge} The unstructured knowledge often contains a great number of tokens, which may lead the decoder to predict the next word directly based on the previously predicted words  without considering the encoder representation. In this case, we reduce the correlation between the generated token $y_t$ and its previous tokens $y_{<t}$ to improve its dependency on the encoder representation $\bm{g}_s$. 

To be specific, we adopt a restricted attention for $y_t$, where $y_t$ merely has access to previous $k$ tokens in the attention mask. Formally, the  unstructured decoder loss for span $s$ is defined as:
\begin{equation}
\begin{aligned}
 &\mathcal{L}_{\text{unstruct}}(Y | s, \theta_{dec}) = \\
 & -\sum_{t}\log P(y_t|y_{t-k:t-1}, \bm{g}_s, \theta_{dec}),
\end{aligned}
\label{eq:unstrucut}
\end{equation} 
where $k$ is the window size of restricted attention.

For the injection process of structured and unstructured knowledge, we use different text markers to obtain span representations from the encoder, respectively, but use a shared decoder $\theta_{dec}$ to process them. Specifically, if a span is associated to multiple pieces of knowledge, we sum up the loss for generating them separately.






\subsection{Overall Training Objective}

In addition to knowledge injection,  we conduct masked language modeling (MLM)~\cite{BERT}  to preserve the model's capabilities in general understanding. MLM samples and replaces 15\% tokens with \texttt{[MASK]} and requires model to predict the missing tokens based on their contextual representations. 

The overall loss of \Ourmodel is composed of three parts: the masked language modeling loss $\mathcal{L}_{\text{MLM}}$, the structured knowledge decoding loss $\mathcal{L}_{\text{struct}}$ and the unstructured knowledge decoding loss $\mathcal{L}_{\text{unstruct}}$. The overall loss can be formulated as:

\begin{equation}
\mathcal{L}= \mathcal{L}_{\text{MLM}}  + \mathcal{L}_{\text{struct}} + \mathcal{L}_{\text{unstruct}}.
\end{equation}

We train the encoder and decoder together during the pre-training process. After that, we discard the decoder and only apply the encoder in downstream applications, which allows our model to be used as easily as the original model without the need for specific plugins. 

\subsection{Training Data Construction}

We combine the Wikipedia annotations and Wikidata facts to construct the pre-training data. We conduct knowledge injection on the anchor span, which contains a hyperlink to a  Wikipedia page and can be associated with a Wikidata entry. 

For the unstructured knowledge injection, we require the decoder to generate the Wikipedia page for each anchor span based on the span representation from the encoder. With the supervision from the decoder, we encourage spans with similar properties, such as the basketball players \emph{Stephen Curry} and \emph{LeBron James}, to obtain similar representations from the encoder.

For the structured knowledge injection, we follow distant supervision~\cite{distant} to align the Wikidata factual knowledge with the Wikipedia text.   
If two entries $h$ and $t$ are in the same context and they have relation $r$ in the Wikidata,  we assume that we can obtain the relational fact $(h, r, t)$ from the context. As shown in Figure~\ref{fig:model}, for the  sentence, {NBA superstar \emph{Stephen Curry} received his Bachelor from \emph{Davidson College}}, we fetch the related fact (Stephen Curry, graduated at Davidson College) from Wikidata. 
After that, we require the decoder to infer the property $(r,t)$ from the span representation of $h$, which help the model to learn the correlation among entity, relation and their context.

Using the above data, we conduct the decoding training objective on anchor spans, which also infuses the linking information into models,  enabling the obtained model to leverage the injected knowledge without extra entity linkers~\cite{tagme}.




\section{Experiments}

In this section, we first introduce the training details of \Ourmodel, followed by the fine-tuning results on the entity-aware tasks and relation-aware tasks.

\subsection{Experimental Setups}

\paragraph{Data Pre-Processing}

We use Wikipedia as our pre-training corpus, which is also adopted in BERT~\cite{BERT}.  We apply the WikiExtractor\footnote{\url{github.com/attardi/wikiextractor}} to extract anchor text with hyperlinks in Wikipedia articles. 
For unstructured knowledge, we use the first up to 64 tokens of the first paragraph of each entity page as the decoder supervision; for structured knowledge, we adopt the relational facts in Wikidata5M~\cite{KEPLER}.  To avoid information leakage for Wiki80~\cite{Wiki80}, we remove the fact that appears in its test set from our training facts.   
In addition, by analyzing the examples constructed by distant supervision, we find that the instance of the top 6 relations occupies about 50\% frequencies, such as the instance of relation \emph{country}, \emph{located in} and \emph{diplomatic relation}. Hence, we remove these 6 frequent but simple relations to accelerate the pre-training process. 


\paragraph{Training Configuration}

Since training \Ourmodel from scratch would be time-consuming, we initialize the parameter of \Ourmodel with \emph{bert-base-uncased}~\cite{BERT} for the main experiments.  
In the subsequent pre-training process, following \citet{RoBERTa}, we remove the next sentence prediction task~\cite{BERT} and train models on contiguous sequences of up to 512 tokens. 
Following \citet{RoBERTa}, we adopt a batch size of 2048 and a peak learning rate of 2e-4 with 10\% warming-up steps. Then, we apply the Adam optimizer~\citep{Adam} to train our model for 10,000 steps. On eight A100 GPUs, the training process for the base model takes 4 hours with fp16 precision and the large model takes 12 hours. The shallow decoder brings about 30\% extra pre-training time compared to the vanilla MLM objective. 
Moreover, the fine-tuning methods and configurations for downstream tasks are presented in the appendix due to space limitations. 

We train three variants of \Ourmodel: \textbf{\Ourmodel (E)}  is only trained with unstructured entity page knowledge; \textbf{\Ourmodel (R)}  is only trained with structured factual knowledge; \textbf{\Ourmodel (E+R)} is trained with both forms of knowledge simultaneously.

\subsection{Baseline}
We compare our models with three kinds of baselines: models without knowledge injection, models with only unstructured knowledge injection, and models with only structured knowledge injection.

Above all, in the absence of knowledge injection, \textbf{BERT}~\cite{BERT} is the original pre-trained model without any further pre-training; \textbf{BERT (our)} continues to pre-train BERT on the same corpus using the same steps as our model. 

Then, as for models merely adopting unstructured knowledge injection, \textbf{TEK}~\cite{TEK} appends unstructured entity page text to the input text until the total length reaches the maximum sequence length, e.g. 512, and then applies BERT (our) to the concatenated input; \textbf{PELT}~\cite{ye2022pelt} constructs embedding of an entity from the sentences in which it appears, then inserts the constructed embedding into the input sequence. 
Moreover, as for the model with only structured knowledge injection, \textbf{ERNIE}~\cite{ERNIE} first encodes the structured factual knowledge as knowledge embeddings~\cite{TransE}, and then integrates knowledge embeddings and original token representations with additional Transformer layers.

Among these baselines, TEK, PELT, and ERNIE require extra entity linkers~\cite{tagme} to find the entity in the text and they also need additional computations to exploit the injected knowledge. Note that all the baselines and our models use less than 100 GPU hours for the further pre-training process.

\subsection{Entity-Aware Task}

We conduct experiments on two entity-aware tasks, entity typing and named entity recognition (NER). Entity typing aims to classify the semantic type of an entity span in a given text.  
NER aims to extract all the named entities' boundaries and their respective types in a given text.

\begin{table}[!t]
\centering
\resizebox{0.48\textwidth}{!}{

\begin{tabular}{l c c c cc }
\toprule
\textbf{Dataset} & \textbf{Task} & \textbf{\#Train} & \textbf{\#Dev}&  \textbf{\#Test} & \textbf{\#Label} \\
\midrule
OpenEntity & {Typing} & 2,000 & 2,000 & 2,000 & 9 \\
\midrule
Few-NERD &  {NER} &  131,767 &    18,824 & 37,648  &   66\\
\midrule
Wiki80 & \multirow{2}{*}{RelC}& 39,200 &5,600 & 11,200 & 80 \\
TACRED &  & 68,124 &22,631 &15,509 &42 \\
\bottomrule

\end{tabular}
}
\caption{Statistics of downstream datasets. Task types include Typing: Entity Typing, NER: Named Entity Recognition, and RelC: Relation Classification. }
\label{tab:data_statistic}
\vspace{-0.5em}
\end{table}
\paragraph{Dataset} 
For the entity typing task, we adopt the manually-annotated OpenEntity dataset~\cite{Openentity}. 
For the NER task, we adopt the Few-NERD dataset~\cite{Fewnerd}, which is a fine-grained NER dataset annotated on Wikipedia sentences.

We evaluate the model in the low-resource setting and the full-data setting. 
We run all experiments with 5 different seeds and report the average score. 
For the  1\% and 10\%  settings, we further randomly select  1\% and 10\% training data from the training set with different seeds for each run.      
As shown in  Table~\ref{tab:data_statistic}, the number of training instances in OpenEntity for 1\% setting is quite small, which is likely to cause unstable results. 
Hence, we only adopt 10\% and 100\%  settings for OpenEntity. 

In addition, to better demonstrate the significance of our improvements, we perform the Student's t-test in the appendix.




\paragraph{Results} As shown in Table~\ref{tab:ent_results}, compared to the models without additional inference cost, \Ourmodel achieves the best performances on both entity-aware datasets. 
With 100\% training data,  \Ourmodel (E) advances BERT (our) by 1.2\% and 0.2\% F1 on OpenEntity and Few-NERD respectively.

By restricting the number of training examples used, we lead PLMs to recall knowledge from their pre-trained parameters rather than re-learn patterns from the training data.  
In the lowest training data setting, \Ourmodel (E) accomplishes more significant results. It outperforms BERT (our) by  15.0\% and 1.3\% F1 on OpenEntity and Few-NERD respectively, which illustrates that the injection of entity page knowledge helps the model to identify entity types.   
Compared to TEK~\cite{TEK}, which directly trusts all retrieved entity pages and appends them to the input text, \Ourmodel (E) learns a global consideration of knowledge during the pre-training phase, thus obtaining more stable improvements.


On the entity-aware tasks, \Ourmodel (E+R)  with both unstructured entity page knowledge and structured factual knowledge injection achieves similar performance to \Ourmodel (E). Though the relational fact cannot further improve model's ability to recognize entity types, different forms of knowledge can be compatibly injected into the encoder using the unified knowledge interface without performance degradation.

\begin{table*}[!t]
\small
\centering
\begin{tabular}{l c |c | cc | ccc}
\toprule
\multirow{2}{*}{\textbf{Model}}  & \multirow{2}{*}{\textbf{Extra Infer Cost}}  & \multirow{2}{*}{\textbf{Inject Type}} & \multicolumn{2}{c|}{\textbf{OpenEntity}}    & \multicolumn{3}{c}{\textbf{Few-NERD}} \\
   &  & &  \textbf{10\%}        & \textbf{100}\%     & \textbf{1\%}        & \textbf{10\%} & \textbf{100\%} \\
\midrule
TEK&\checkmark& U & 42.9$_{\pm {5.7}}$ & 74.1$_{\pm {0.5}}$ & 57.8$_{\pm {0.2}}$ & 63.2$_{\pm {0.3}}$ &  67.7$_{\pm {0.1}}$\\
PELT &\checkmark& U & 46.9$_{\pm {3.9}}$ & 74.0$_{\pm {0.3}}$ & 57.8$_{\pm {0.2}}$ & 63.0$_{\pm {0.2}}$ &  67.7$_{\pm {0.1}}$\\
 ERNIE &\checkmark& S &  53.8$_{\pm {7.9}}$ &  74.7$_{\pm {0.3}}$ & \femph{59.3$_{\pm {0.5}}$} &  \femph{63.7$_{\pm {0.3}}$} & {67.8$_{\pm {0.1}}$}  \\
 \midrule

BERT & -&- & 49.0$_{\pm {5.9}}$ &  74.7$_{\pm {0.3}}$ & 57.2$_{\pm {0.5}}$  & 63.0$_{\pm {0.3}}$ &  67.6$_{\pm {0.1}}$ \\
BERT (our)&- & - & 49.1$_{\pm {3.4}}$ & 74.2$_{\pm {0.4}}$ & 57.8$_{\pm {0.2}}$ & 63.2$_{\pm {0.3}}$ &  67.7$_{\pm {0.1}}$\\

 \Ourmodel (E) & -&U & \femph{{64.1}$_{\pm {1.7}}$}  & \semph{75.4}$_{\pm {0.3}}$  & {58.5$_{\pm {0.3}}$} & \semph{63.5$_{\pm {0.1}}$} & \semph{67.9$_{\pm {0.1}}$}\\
 
 \Ourmodel (R) &  -&S & {63.0$_{\pm {1.9}}$}  & \femph{{75.5}}$_{\pm {0.6}}$   & {58.4$_{\pm {0.4}}$} & 63.3$_{\pm {0.2}}$ &67.8$_{\pm {0.1}}$    \\
 
 \Ourmodel (E+R) & -& U+S & \semph{63.4}$_{\pm {1.6}}$& {74.9$_{\pm {0.2}}$}  & \semph{58.9$_{\pm {0.3}}$}& {63.4$_{\pm {0.2}}$}& \femph{68.0$_{\pm {0.1}}$} \\
\bottomrule
\end{tabular}
\caption{Results on entity-aware tasks.  x\% indicate using x\% supervised training data.  Injected knowledge includes  \textbf{U}: Unstructured knowledge and \textbf{S}: Structured knowledge.  
We report the average micro-F1 across 5 runs with the standard deviation as the subscript. We color the best result with \colorbox[RGB]{197,224,180}{green} and color the second best result with \colorbox[RGB]{207,226,243}{blue}. }
\label{tab:ent_results}
\end{table*}

\begin{table*}[!t]
\small
\centering
\begin{tabular}{ l c | c | ccc| ccc}
\toprule

\multirow{2}{*}{\textbf{Model}}  & \multirow{2}{*}{\textbf{Ex. Infer Cost}}  & \multirow{2}{*}{\textbf{Inject Type}}  &  \multicolumn{3}{c|}{\textbf{Wiki80}}   & \multicolumn{3}{c}{\textbf{TACRED}} \\
     &&           & \textbf{1\%}        & \textbf{10\%} & \textbf{100\%}   & \textbf{1\%}        & \textbf{10\%} & \textbf{100\%}  \\
\midrule
TEK &\checkmark&  U & 43.3$_{\pm {0.7}}$ & 80.1$_{\pm {0.7}}$ & \femph{92.1}$_{\pm {0.1}}$  & 19.1$_{\pm {3.6}}$  & 55.7$_{\pm {0.9}}$ & 67.0$_{\pm {0.7}}$  \\
PELT &\checkmark&  U & 46.6$_{\pm {0.9}}$ & 79.3$_{\pm {0.8}}$ & {91.4}$_{\pm {0.3}}$  & 21.9$_{\pm {4.0}}$  & 56.4$_{\pm {0.7}}$ & 67.6$_{\pm {0.5}}$  \\
ERNIE &\checkmark&  S  & 48.8$_{\pm {1.5}}$ &  \semph{84.1$_{\pm {0.4}}$} &91.1$_{\pm {0.2}}$ &20.0$_{\pm {4.5}}$ & 56.0$_{\pm {0.7}}$ & 67.0$_{\pm {0.7}}$ \\
\midrule
BERT &- &- & 45.5$_{\pm {1.4}}$& 77.5$_{\pm {0.5}}$ & 90.9$_{\pm {0.1}}$  & 21.3$_{\pm {4.0}}$  & 55.5$_{\pm {0.6}}$ & 67.1$_{\pm {0.5}}$ \\
BERT (our) &- & - & 46.2$_{\pm {1.6}}$ & 78.5$_{\pm {0.4}}$ &90.9$_{\pm {0.2}}$  & 21.4$_{\pm {3.6}}$  & 57.4$_{\pm {0.9}}$ & 67.9$_{\pm {0.4}}$  \\
\Ourmodel (E) & - &U &{50.5$_{\pm {1.6}}$}  &82.5$_{\pm {0.3}}$  & {91.4$_{\pm {0.2}}$}& \semph{29.6$_{\pm {3.7}}$} &{58.6$_{\pm {0.5}}$} & {68.0$_{\pm {0.4}}$}\\
\Ourmodel (R) & - &S & \semph{52.0$_{\pm {1.7}}$}  &83.7$_{\pm {0.6}}$  & 91.6$_{\pm {0.2}}$ & {24.1$_{\pm {2.8}}$} &\semph{58.7$_{\pm {0.8}}$} & \femph{68.3$_{\pm {0.6}}$}\\

 \Ourmodel (E+R) & - &U+S & \femph{58.2$_{\pm {1.7}}$}   &  \femph{86.6$_{\pm {0.1}}$}& \femph{92.1$_{\pm {0.0}}$}& \femph{33.6$_{\pm {2.6}}$} &\femph{59.7$_{\pm {0.7}}$} & \semph{68.2$_{\pm {0.2}}$}\\
\bottomrule
\end{tabular}

\caption{Results on relation-aware tasks. Injected knowledge includes \textbf{U}: Unstructured knowledge and \textbf{S}: Structured knowledge. 
We report accuracy on Wiki80 and micro-F1 on TACRED across 5 runs with the standard deviation as the subscript.  We color the best result with \colorbox[RGB]{197,224,180}{green} and color the second best result with \colorbox[RGB]{207,226,243}{blue}.} 
\label{tab:rel_results}
\vspace{-0.5em}
\end{table*}

\subsection{Relation-Aware Task}

We adopt the relation classification task for evaluation. Given a subject entity and an object entity,  relation classification aims to predict the relationship between them in context. 

\paragraph{Dataset} We adopt two representative relation classification datasets, Wiki80~\cite{Wiki80} and TACRED~\cite{TACRED}. Wiki80 contains 80 relations and each relation has 700 instances from Wikipedia. TACRED  contains 42 relations in TAC KBP corpus, which, different from Wiki80, includes a special relation \emph{no\_relation}. 
We use the TAGME tool~\cite{tagme} to annotate  Wiki80 rather than using the golden annotation~\cite{ERNIE}  for a fair comparison. 
Similar to the evaluation on entity-aware task, we also assess models in  the low-resource setting and the full-data setting as well as perform the Student's t-test in the appendix.


\paragraph{Results}  
As shown in Table~\ref{tab:rel_results},  on the Wiki80 of the Wikipedia domain,   \Ourmodel (E) boosts BERT (our) by 4.3\%, 4\%, and 0.5\% F1 in the 1\%, 10\%, and 100\% settings respectively, which indicates that the injected entity page knowledge includes a large amount of factual knowledge.  \Ourmodel (R) improves BERT (our) by 5.8\%, 5.2\%, and 0.7\% F1 in the 1\%, 10\%, and 100\% settings, which demonstrates the effectiveness of injecting factual knowledge directly into the model for the relation classification task. Moreover,  \Ourmodel (E+R) has been shown to integrate unstructured and structured knowledge through a unified interface and achieves the best performance in all scenarios.

While the baseline ERNIE~\cite{ERNIE} augmented with factual knowledge from the Wikipedia domain shows little improvement over out-of-domain TACRED, \Ourmodel (R) learns relational patterns from the distantly aligned examples, achieving continuous improvements on the TACRED dataset.  Moreover, with two forms of knowledge injection, \Ourmodel (E+R)  obtains the best +11.2\% and +2.3\% F1 improvements over BERT (our) on TACRED in 1\% and 10\% settings respectively. It demonstrates the importance of extending the diversity of knowledge injection to handle examples in various domains.


\section{Discussion}

\begin{table*}[!t]
\small
\centering
\begin{tabular}{ l | cc|ccc| ccc|ccc}
\toprule

\multirow{2}{*}{\textbf{Model}}   & \multicolumn{2}{c|}{\textbf{OpenEntity}} &  \multicolumn{3}{c|}{\textbf{Few-NERD}} &  \multicolumn{3}{c|}{\textbf{Wiki80}}  &
\multicolumn{3}{c}{\textbf{TACRED}} \\
     &     10\%    &  100\%    & 1\%& 10\% & 100\%         & 1\%&    10\% &   100\%   &1\%& 10\%   & 100\%     \\
\midrule
BERT\blarge(our) & 47.6 & 75.1  &59.6&  64.1&68.1& 46.3&79.5&92.2 &18.7 &55.7&64.2 \\
 UNTER\blarge(E+R) & \textbf{63.5}&\textbf{75.3} &\textbf{59.8}&\textbf{65.9}&\textbf{68.2}&\textbf{64.6}& \textbf{89.2}&\textbf{93.1} &\textbf{35.7} &\textbf{61.8}&\textbf{70.0}\\
 \midrule
 RoBERTa\blarge(our) & 59.6 & 75.5   & 62.1& 64.2& 69.0 & 54.2&80.3&85.6 & 23.8 &57.7&65.4 \\
 Ro-UNTER\blarge(E+R) & \textbf{65.5}&\textbf{76.0} & \textbf{62.8}&\textbf{66.0}& \textbf{69.4}&\textbf{66.7}& \textbf{90.2}&\textbf{93.4} & \textbf{39.0} &\textbf{64.0}&\textbf{71.8}  \\
\bottomrule
\end{tabular}

\caption{Results on large models. x\%  indicate using  x\% supervised training data.  
 We report the average score across 5 runs.   } 
\label{tab:large_results}
\vspace{-0.5em}

\end{table*}


\begin{table}[!t]
\small
\centering
\begin{tabular}{l ccc}
\toprule
\textbf{Att. Width} & \textbf{OpenEntity} & \textbf{Wiki80} & \textbf{TACRED} \\
\midrule
2 & \textbf{64.1}$_{\pm {1.7}}$ & \textbf{82.5}$_{\pm {0.3}}$ & \textbf{58.6}$_{\pm {0.5}}$\\
4 & 63.4$_{\pm {1.4}}$ & {82.3}$_{\pm {0.3}}$ & {58.5}$_{\pm {0.8}}$\\
16 & 62.7$_{\pm {2.2}}$ & 82.3$_{\pm {0.3}}$ & 58.0$_{\pm {0.6}}$\\
All &  62.5$_{\pm {2.0}}$  & 82.1$_{\pm {0.4}}$ & 57.6$_{\pm {0.7}}$\\
\bottomrule
\end{tabular}
\caption{Results on  \Ourmodel (E) with different decoder attention span widths. We adopt 10\%  setting for experiments.} 
\label{tab:capacity_attention_results}
\end{table}

\begin{table}[!t]
\small
\centering
\begin{tabular}{l ccc}
\toprule
\textbf{Decoder Layer} & \textbf{OpenEntity} & \textbf{Wiki80} & \textbf{TACRED} \\
\midrule
1 & {63.4}$_{\pm {1.6}}$ & \textbf{86.6}$_{\pm {0.1}}$ & \textbf{59.7}$_{\pm {0.7}}$\\
2 & \textbf{63.9}$_{\pm {1.9}}$ & {86.4}$_{\pm {0.3}}$ & {59.3}$_{\pm {0.9}}$\\
3 & \multicolumn{3}{c}{OverFiting in Pre-training}\\
\bottomrule
\end{tabular}
\caption{Results on  \Ourmodel (E+R)  with different decoder layers. We adopt 10\%  setting for experiments.} 
\label{tab:capacity_layer_results}
\end{table}

\subsection{How the Capacity of the Decoder Affects?}
 In this section, we investigate how the capacity of the decoder affects the downstream performance of the pre-trained encoder. Above all, to quickly recap, in our proposed framework, the decoder is used as a unified knowledge interface and its capacity is limited so that we can discard it after the pre-training process. Specifically, We restrict the capacity of decoder from two perspectives, the attention span width of generated tokens and the number of decoder layers. 
We conduct experiments with 10\% setting on OpenEntity, Wiki80, and TACRED.

As shown in Table~\ref{tab:capacity_attention_results}, \Ourmodel (E) achieves the best performance with attention span width of 2, while the model with sequential full attention performs the worst. Hence, we choose attention span width as 2 for all models in main experiments.

As shown in Table~\ref{tab:capacity_layer_results}, \Ourmodel (E+R) with one decoder layer performs slightly better on the relation classification task than the model with two decoder layers, but it performs worse on entity typing.  When the number of decoder layers reaches 3, the model overfits the distantly-constructed factual knowledge, resulting in a loss that exceeds the bounds of fp16 precision. Since one-layer decoder runs faster than the two-layer one, we choose one-layer decoder in main experiments.

Overall, in the knowledge injection process, the stronger the decoder, the weaker the encoder. Hence, we restrict the capacity of the decoder to enable the model to store as much knowledge as possible into the encoder.

\subsection{Which  Span Representation is Better?}

\begin{table}[!t]
\small
\centering
\begin{tabular}{l ccc}
\toprule
\textbf{Model} & \textbf{OpenEntity} & \textbf{Wiki80} & \textbf{TACRED} \\
\midrule
BERT (our) & 49.1$_{\pm {3.4}}$ & 78.5$_{\pm {0.4}}$& 57.4$_{\pm {0.9}}$ \\ 
Token-Concat & 47.0$_{\pm {2.9}}$ & 78.1$_{\pm {0.4}}$&56.5$_{\pm {0.8}}$ \\
Marker & \textbf{63.4}$_{\pm {1.6}}$ & \textbf{86.6}$_{\pm {0.1}}$& \textbf{59.7}$_{\pm {0.2}}$ \\
\bottomrule
\end{tabular}
\caption{Results with different encoder span representations. We adopt 10\%  setting for experiments.} 
\label{tab:spanrep_results}
\vspace{-0.5em}
\end{table}

We supervise the encoder span representation via the signal from the decoder. To obtain the span representation, a straightforward approach is to concatenating the representations of span's boundary tokens as the span representation~\cite{spanbert,corefbert}, denoted as \textbf{Token-Concat}.  In this work, instead of using tokens' final representations,  we leverage levitated markers to aggregate span's information across Transformer layers to acquire the span representation, denoted as \textbf{Marker}.  

We compare the Token-Concat span representation and the marker-based span representation on \Ourmodel (E+R). As shown in Table~\ref{tab:spanrep_results}, the marker-based span representation enables the enhanced model to achieve the best result on OpenEntity, Wiki80, and TACRED. In contrast, knowledge injection with Token-Concat span representations even weaken the model performance. It illustrates that applying knowledge injection loss on the token representation would interfere with token's original masked language modeling. Hence, we use additional marker-based span representation to decouple the knowledge injection and the masked language modeling.

\subsection{Is \Ourmodel  Effective for Larger Model?}

To  validate the effectiveness of our injection method with larger models, we train a series of models as follows.  
\textbf{UNTER\blarge(E+R)} is initialized with the weight of \emph{bert-large-uncased}~\cite{BERT} and then further pre-trained with the MLM objective as well as structured and unstructured knowledge injection objectives for 10,000 steps.  Toward a fair comparison, we follow the same configuration to obtain \textbf{BERT\blarge(our)} with only the MLM objective.  Similarly, we acquire \textbf{Ro-UNTER\blarge(E+R)} and \textbf{RoBERTa\blarge(our)} from the initialization of  \emph{roberta-large}~\cite{RoBERTa}.

As shown in Table~\ref{tab:large_results}, \Ourmodel achieves continuous improvements over BERT\blarge(our) and RoBERTa\blarge(our). It illustrates that our knowledge injection method can be generalized to models of different sizes. Specifically, in the 100\% setting of Wiki80 in the Wiki domain,  RoBERTa\blarge(our) performs worse than BERT\blarge(our), because the latter is pre-trained with a higher proportion of the Wiki-domain corpus. Moreover, trained with our knowledge injection, Ro-UNTER\blarge(E+R) can be adapted to the Wiki domain more faster than the model further pre-trained with only the MLM objective~\cite{dontstoppretraining}. 

\section{Conclusion}
In this work,  we propose a unified knowledge interface to integrate the injection of different forms of knowledge, which enables the enhanced encoder to exploit the injected knowledge without specific plugins. Experimental results show that with both unstructured and structured knowledge injected, our model can significantly improve PLM's performance on a range of knowledge-driven NLP tasks,  especially in  low-resource scenarios. 

\section*{Limitation}
A limitation of this work is that we only validate our unified knowledge interface in terms of enhancing encoder models. However, the generation models, such as autoregressive models~\cite{GPT-2} and encoder-decoder models~\cite{bart,t5}, also play an important role in many NLP tasks. In the future, we will explore more ways to unify the injection of different forms of knowledge into PLMs with different architectures. 



\bibliography{anthology,custom}
\bibliographystyle{acl_natbib}

\appendix
\newpage

\section{Fine-tuning Details}

The fine-tuning methods for different task:

\paragraph{Entity Typing} Following ERNIE~\cite{ERNIE}, we first insert a pair of special tokens to enclose and emphasize the given span in the input sequence. After PLM's encoding, we adopt the final representation of the [CLS] token for classification.

\paragraph{Named Entity Recognition} Following ~\citet{BERT}, we regard the  task as a sequence labeling task and apply a token-level classifier to tokens' contextual representations to predict their IOB2 tags~\cite{BIO}. 

\paragraph{Relation Extraction} Following ERNIE~\cite{ERNIE}, we insert two pairs of special tokens to enclose and highlight the subject entity and the object entity in the input sequence respectively. After that, we add a relation classifier layer on the [CLS] token's final representation for predictions.

We fine-tune the downstream models using Adam optimizer~\cite{Adam} with 10\% warming-up steps. We list our hyper-parameters in Table~\ref{tab:Hyperparameters}. 
We run all experiments with 5 different seeds (42, 43, 44, 45, 46). In experiments with 1\% and 10\% settings, we randomly remain 1\% and 10\% training data respectively with 5 different seeds (42, 43, 44, 45, 46) for 5 runs. Moreover, we adopt 10-50 times the epochs of the experiments with the 100\% setting when training the model with 1\% and 10\% settings


\begin{table}[ht]
\centering
\small
\begin{tabular}{l | c c c } 
\toprule
\textbf{Dataset} &  \textbf{Epoch}   & \textbf{Batch Size} & \textbf{Learning Rate} \\
\midrule
OpenEntity & 10 & 16 & 2e-5 \\
Few-NERD &3 & 8 & 2e-5\\
Wiki80 &5 & 32 & 3e-5\\
TACRED &5 & 32 & 2e-5\\
Wiki-ET &3 & 32 & 2e-5\\
\bottomrule
\end{tabular}
\caption{Hyper-parameters for fine-tuning. }
\label{tab:Hyperparameters}
\end{table}



\section{Results on Wiki-ET}
Beside OpenEntity~\cite{Openentity}, we also evaluate models on another entity typing dataset, Wiki-ET~\cite{wikiet}, which is automatically built by aligning Wikipedia and Freebase. Wiki-ET is extremely large but also contains much noise. As shown in Table~\ref{tab:wikiet}, \Ourmodel (E+R) beat baselines with different backbone models, including BERT$_{\small \texttt{B}}$, BERT\blarge and RoBERTa\blarge, which again shows the effectiveness of our knowledge injection in the entity typing task. Notably, due to the noise in dataset, many models achieves better performance with 10\% training data than that those with 100\% training data.

\begin{table}[!t]
\centering
\small
\begin{tabular}{l | c c c } 
\toprule
\textbf{Model} &  \textbf{1\%}   &  \textbf{10\%} & \textbf{100\%}\\
\midrule
BERT (our) & 76.4$_{\pm {0.1}}$ &  77.8$_{\pm {0.3}}$ &  77.8$_{\pm {0.5}}$ \\
\Ourmodel (E+R) & \textbf{77.6}$_{\pm {0.5}}$ &  \textbf{78.6}$_{\pm {0.3}}$ &  \textbf{78.2}$_{\pm {0.3}}$ \\
\midrule
BERT\blarge(our) &77.4$_{\pm {0.7}}$ &77.7$_{\pm {0.5}}$&  78.0$_{\pm {0.4}}$ \\
UNTER\blarge(E+R) & \textbf{78.4}$_{\pm {0.6}}$ &\textbf{79.0}$_{\pm {0.2}}$ &\textbf{78.5}$_{\pm {0.6}}$ \\
\midrule
RoBERTa\blarge(our) & 79.2$_{\pm {0.6}}$ &79.4$_{\pm {0.3}}$ &78.7$_{\pm {0.5}}$ \\
Ro-UNTER\blarge(E+R) & \textbf{79.7}$_{\pm {0.4}}$ &\textbf{80.1}$_{\pm {0.4}}$&\textbf{79.7}$_{\pm {0.4}}$ \\
\bottomrule
\end{tabular}
\caption{Results on the Wiki-ET dataset using different proportions of training data. }
\label{tab:wikiet}
\end{table}

\section{Significance Test}
In the paper, we run all experiments 5 times and report the average score. As shown in Table~\ref{tab:ent_results} and Table~\ref{tab:rel_results}, UNTER (E+R) model beats BERT (our) baseline by 1.7-14.4 F1 in low-resource settings and outperforms the baseline by 0.3-1.2 F1 in the full-data setting. 
Using numbers from 5 runs of the experiment, we perform Student's t-test:
\begin{align*}
  H_0: \mu_{\text{baseline}} &>= \mu_{\text{UNTER (E+R)}} \\
  H_1: \mu_{\text{baseline}} &<\ \ \, \mu_{\text{UNTER (E+R)}}
\end{align*}

As shown in Table~\ref{tab:t_test}, for almost settings, we reject H0 and accept H1 at a significance level of 0.005. It demonstrate that our unified knowledge injection can significantly improve PLMs on the evaluation tasks.

\begin{table}[ht]
\centering
\small
\begin{tabular}{ l | ccc}
\toprule
\multicolumn{4}{c}{\textbf{OpenEntity}} \\
\midrule
\textbf{Setting} &  \textbf{1\%} &  \textbf{10\%}    &  \textbf{100\%} \\
\midrule
t-statistic &-& 8.5 & 3.5 \\
Significant Level $\alpha$ & -& < 0.005&< 0.005 \\
\midrule

  \multicolumn{4}{c}{\textbf{Few-NERD}} \\

 \midrule
 t-statistic  & 6.8 & 1.2 & 4.7 \\
 Significant Level $\alpha$ &  < 0.005 & < 0.1 &< 0.005\\
\midrule
  \multicolumn{4}{c}{\textbf{Wiki80}}  \\
 \midrule
t-statistic & 11.5 & 55.9 & 13.4 \\
Significant Level $\alpha$& < 0.005&< 0.005&< 0.005 \\
\midrule
 \multicolumn{4}{c}{\textbf{TACRED}} \\
\midrule
t-statistic &  6.1 & 4.5 & 1.5 \\
Significant Level $\alpha$ &  < 0.005 &   < 0.005 & < 0.1 \\
\bottomrule
\end{tabular}
\caption{t-statistic and corresponding significance level to reject  $H_0$ and accept $H_1$.}
\label{tab:t_test}
\end{table}


\end{document}